\documentclass{article}

\usepackage{arxiv}
\usepackage{authblk}
\usepackage[utf8]{inputenc}
\usepackage[T1]{fontenc}
\usepackage{hyperref}
\usepackage{url}
\usepackage[numbers]{natbib}  
\usepackage{booktabs}
\usepackage{amsfonts}
\usepackage{nicefrac}
\usepackage{microtype}
\usepackage{graphicx}
\usepackage{xcolor}
\graphicspath{{./image/}}

\usepackage{bm}
\usepackage{amsmath,amssymb}
\usepackage{amsthm}
\usepackage[shortlabels]{enumitem}
\usepackage{siunitx}
\usepackage{listings}
\usepackage{float}
\usepackage{placeins}
\usepackage{multirow} 
\usepackage[ruled,vlined]{algorithm2e}

\theoremstyle{definition}

\title{Exploring Causes and Mitigation of Hallucinations in Large Vision Language Models}

\author{Yaqi Sun}
\author{Kyohei Atarashi}
\author{Koh Takeuchi}
\author{Hisashi Kashima}

\affil{Kyoto University}

\begin{document}
\maketitle

\begin{abstract}
Large Vision Language Models (LVLMs) integrate image encoders with Large Language Models (LLMs) to process multi-modal inputs and perform complex visual tasks. However, they often generate hallucinations by describing non-existent objects or attributes, limiting their reliability. This study analyzes hallucination patterns in image captioning, showing that not all tokens in the generation process are influenced by image input and that the dependency on image input can be helpful to detect hallucinations. To address this, we develop an automated pipeline to identify hallucinated objects and train a token-level classifier using hidden representations from parallel inference passes—with and without image input. Leveraging this classifier, we introduce a decoding strategy that effectively controls hallucination rates in image captioning at inference time.
\end{abstract}

\section{Introduction}\label{sec-intro}

\subsection{Background}\label{subsec-back}
Computer vision has undergone remarkable advancement over the past decade. The field evolved from Convolutional Neural Networks (CNNs) to Vision Transformers (ViTs), achieving significant progress in various visual tasks. Notably, the Contrastive Language-Image Pre-training (CLIP) framework~\cite{radford2021learning} provided technical foundations for robust vision-language alignment by mapping visual and textual inputs into a shared embedding space through contrastive learning.

On the other hand, Large Language Models (LLMs), powered by transformer architectures and trained on vast text corpora for next-token prediction, have demonstrated exceptional capabilities in natural language tasks. These advances in both vision-language alignment and LLMs have catalyzed the development of Large Vision Language Models (LVLMs).

However, in basic vision tasks such as object detection and image classification, traditional task-specific approaches consistently achieve superior performance compared to current LVLMs~\cite{zhang2024visually}. Current LVLMs, including proprietary models such as GPT-4V~\cite{achiam2023gpt}, Gemini~\cite{team2023gemini}, and open-source alternatives like LLaVA~\cite{liu2024visual,liu2024improved}, are known to share similar architectural designs and training paradigms. While these models show promising multi-modal understanding abilities, they frequently produce hallucinations, which presents significant challenges for the practical deployment of LVLMs. As multi-modal alignment becomes increasingly crucial for artificial general intelligence, the research community has focused on reducing hallucinations and developing robust evaluation methods.

\subsection{Overview}\label{subsec-overview}
Although LVLMs have inherited powerful language capabilities from their text-only base model, they exhibit distinct hallucination patterns in multi-modal tasks. In this study, we focus on the hallucination problem of LVLMs in the image captioning task. In this context, hallucination refers to instances where models generate descriptions containing objects, attributes, or spatial relationships that mismatch the input image.

Our experimental analysis reveals concerning patterns in the generation process: first, hallucinations are more likely to occur in the later part of the caption. Second, models exhibit inconsistent object recognition between generative and discriminative tasks. Moreover, when the image input is removed, the token probability distributions remain largely unchanged throughout the generation process. These observations suggest that the model gradually reduces its dependence on visual information, increasingly relying on language priors during open-ended generation.

Inspired by these observations, we propose an approach that detects and reduces hallucinations by comparing token-level hidden representations between two inference passes—one with the image and one without. Our method does not require human annotation or model fine-tuning, instead utilizing a light-weight classifier trained on automatically labeled data to guide inference.

Our study introduces three contributions:
\begin{itemize}
    \item \textbf{Automated annotation pipeline:} We developed an automated annotation pipeline that integrates multiple open-vocabulary object detection tools to identify hallucinated content in image captions.
    \item \textbf{Token-level hallucination classifier:} A token-level hallucination classifier is trained on hidden representations, enabling the detection of potential hallucinations during the generation process.
    \item \textbf{Novel decoding method:} We propose a novel decoding method that integrates the hallucination classifier and sampling to control hallucination rates during inference, significantly reducing hallucinated content.
\end{itemize}

\section{Related work}\label{sec-relatedwork}

\subsection{Architecture and training of LVLMs}\label{subsec-training}
Large Vision Language Models (LVLMs) have commonly adopted an architecture that connects pre-trained vision and language backbones through a lightweight projector. In this architecture, the visual encoder extracts high-dimensional visual tokens (often using a CLIP-based model), which are then projected into the language model's embedding space for unified processing. Notable examples following this approach include BLIP-2~\cite{li2023blip}, InstructBLIP~\cite{dai2023instructblip}, LLaVA~\cite{liu2024visual, liu2024improved}, and MiniGPT-4~\cite{zhu2023minigpt}.

Recent studies suggest that hallucination patterns in LVLM outputs are model-specific and closely related to both architectural choices and the quality and scale of the training data~\cite{liu2024survey, gunjal2024detecting}. Since LLaVA-1.5-7B~\cite{liu2024improved} embodies current mainstream practices in LVLM development, including a fully auto-regressive architecture and a representative training methodology, we adopt it as our base model in this study.

The model consists of three key components: CLIP ViT-L/14~\cite{radford2021learning} as the vision encoder, Vicuna-7B~\cite{vicuna2023} as the language model, and a multi-layer perceptron (MLP) projector. These components are trained through a two-stage process:

\begin{itemize}
    \item \textbf{Stage 1 (pre-training):} The projector is trained on 595K image-text pairs to align visual and textual embeddings.
    \item \textbf{Stage 2 (fine-tuning):} The model is fine-tuned on 158K instruction-following data and ScienceQA, a multi-modal scientific QA dataset. Task-oriented samples such as image captioning and visual question answering (VQA) are used to enhance its instruction-following capabilities.
\end{itemize}

Both the pre-training and fine-tuning datasets are relatively small, making the training process efficient. However, this limited size may affect the model's ability to generalize well across diverse tasks.

\subsection{Mitigating hallucinations in LVLMs}\label{subsec-mitigating}
Methods to mitigate hallucinations in LVLMs can be broadly categorized into training-based and training-free methods. 

Training-based methods typically involve collecting fine-grained feedback data, either through manual annotation or LLM-assisted recaptioning frameworks. This data is then used to fine-tune the model using Reinforcement Learning from Human Feedback (RLHF), which can effectively reduce hallucinations but often comes with high costs in data collection and model training. 

Training-free methods do not involve fine-tuning the LVLM and can be divided into two categories: inference-time methods and post-hoc frameworks. OPERA and VCD represent inference-time methods that modify token distributions during generation, while LURE and Woodpecker exemplify post-hoc frameworks that either train a revisor using hallucinated data or employ visual grounding tools for output validation.

\subsubsection{Inference-time methods}\label{subsubsec-inference}
\textbf{Over-trust Penalty with Retrospection-Allocation (OPERA)}~\cite{huang2024opera} is an inference-time method designed to address the issue of models over-relying on specific \textquotedblleft anchor tokens\textquotedblright{} during generation, which often leads to hallucinations. It first applies a penalty score based on local attention aggregation patterns. During beam search, this penalty dynamically adjusts token selection, encouraging a more balanced attention distribution across image and text tokens. In cases where the over-trust penalty fails to prevent aggregation, retrospection-allocation strategy actively redirects generation by selecting alternative tokens when the model exhibits repetitive focus on specific anchor tokens.

\textbf{Visual Contrastive Decoding (VCD)}~\cite{leng2024mitigating} mitigates object hallucinations by addressing training data biases and over-reliance on language priors. The method contrasts outputs from original and noise-distorted visual inputs to adjust the model's output distribution. By favoring high-confidence tokens from original inputs, VCD generates more accurate outputs while maintaining efficiency.

Although OPERA and VCD are effective decoding methods, both approaches require extensive manual parameter tuning. OPERA needs predetermined penalty coefficients that are difficult to optimize across different scenarios, and VCD requires careful calibration of token-specific contrast parameters to balance hallucination rate and fluency. Our approach addresses these limitations through a supervised token-level classifier that automatically learns model-specific hallucination patterns, eliminating the need for manual parameter tuning and providing intuitive control over hallucination rates during decoding.

\subsubsection{Post-hoc frameworks}\label{subsubsec-posthoc}
\textbf{LVLM Hallucination Revisor (LURE)} ~\cite{zhou2023analyzing} rectifies object hallucinations in LVLMs by addressing three primary causes of hallucination: co-occurrence biases in training data, uncertainty in object predictions, and errors accounting later parts of generated descriptions. To train the revisor, LURE generates hallucinatory training data by modifying accurate captions using GPT-3.5, introducing co-occurring objects and replacing uncertain or late-positioned objects with placeholder tags. The revisor model is trained on this dataset to learn how to correct hallucinatory outputs into the accurate description. Once trained, the revisor can seamlessly integrate with any LVLM to correct hallucinations. 

\textbf{Woodpecker}~\cite{yin2024woodpecker} identifies key concepts in generated text and formulates diagnostic questions using prompt engineering to detect inconsistencies, covering both object-level and attribute-level details. These diagnostic questions are answered using object detection tools and bounding boxes, which validate image content and produce structured visual claims. The visual claims are then used to guide the correction process, refining the generated text to better align with the image. This structured and interpretable approach improves the accuracy and reliability of the LVLM output.

Compared with post-hoc frameworks such as LURE and Woodpecker, our approach avoids the need for manual hallucination pattern identification or designing hallucinated datasets. Instead, we employ a model-specific classifier that automatically learns these patterns. Our method operates independently during decoding without relying on external tools for caption validation, enabling real-time control. However, note that integrating our method with other LVLMs requires retraining of the classifier.

\subsection{Evaluating hallucinations in LVLMs}\label{subsec-evaluating}
Recent work has also developed various approaches to evaluate hallucinations in LVLMs through discriminative assessment methods. \textbf{Polling-based Object Probing Evaluation (POPE)}\cite{li2023evaluating} implements a binary classification framework where models provide "Yes/No" responses to object existence queries. This approach avoids complex parsing of generated captions and ensures stable and fair evaluations. \textbf{Recognition-based Object Probing Evaluation (ROPE)}\cite{chen2024multi} addresses multi-object hallucinations using bounding box prompts, examining factors such as object salience and class distribution that influence hallucination occurrence. They note that decoding strategies such as OPERA show limited effectiveness in diverse multi-object scenarios, sometimes even lowering performance.

\textbf{AMBER}~\cite{wang2023llm} proposes to evaluate generative and discriminative tasks as two distinct aspects of LVLM's performance. The method assesses discriminative performance through structured prompts that test different hallucination types. Additionally, it evaluates generative performance by comparing model outputs with ground-truth annotations of objects, attributes, and relationships.

Free-form text evaluation presents an essential challenge in hallucination detection. The \textbf{Caption Hallucination Assessment with Image Relevance (CHAIR)}\cite{rohrbach2018object} metric evaluates object hallucinations but requires object segmentation annotations and semantic parsing to map detected nouns to a predefined object vocabulary. \textbf{HalluBench}\cite{zhao2023beyond} addresses these constraints using GPT-4 to evaluate hallucinations against detailed descriptions at the object level, allowing evaluation of objects, attributes and relationships.

The dependence on ground-truth annotations and the challenge in free-form text processing create a shared bottleneck for both evaluation and mitigation methods. The ability to evaluate hallucination levels in captions without reference data would facilitate the reduction of hallucinations in image captioning, thus enabling the collection of high-quality hallucination-free training data.

\section{Problem setting}\label{sec-problemsetting}

\subsection{Task definition}\label{subsec-task}
Given a single input image \textit{I}, we employ LLaVA-1.5-7B to generate a descriptive caption \textit{C} using the prompt "Describe the image in detail." This constitutes an open-ended generative task where the output format is unrestricted, and the maximum sequence length is set to 512 tokens.

Our study focuses on object hallucination in this image captioning setting. Our primary objective is to develop a decoding strategy that mitigates object hallucination in the generated captions at inference time.

\subsection{Hallucination definition}\label{subsec-hallucination}
In the context of image captioning, hallucination refers to the mismatch between the generated caption and the visual content. We can categorize hallucinations as follows:

\begin{enumerate}
    \item \textbf{Object hallucination}: Inconsistency in object presence between the caption and image. This refers to objects mentioned in caption \( C \) but absent in image \( I \). While objects present in \( I \) but omitted in \( C \) could also be considered a form of hallucination, this aspect is less emphasized in current research.
    
    \item \textbf{Attribute hallucination}: Incorrect description of object properties, including: 
    \begin{itemize}
        \item Quantities (e.g., number of objects)
        \item Physical properties (e.g., color, shape, size, state)
        \item Spatial relationships (e.g., \textquotedblleft person near the door\textquotedblright)
    \end{itemize}
    
    \item \textbf{Contextual hallucination}: Unsupported inferences about the scene, including perceived emotions, ongoing actions, or environmental details that are not directly evident in the image.
\end{enumerate}

\section{Analyzing hallucination patterns in LVLMs}\label{sec-analyzingpatterns}
Our experiments reveal several distinct patterns in how Large Vision Language Models (LVLMs) generate hallucinations, providing insight into their underlying mechanisms and potential mitigation strategies. 

Section~\ref{subsec-patterns} reveals trends in hallucination frequency related to token positions, confirming phenomena previously observed in LURE~\cite{zhong2024investigating}. Section~\ref{subsec-pipeline} presents a novel pipeline that combines object detection tools to annotate object hallucination in image captions, enabling a scalable detection framework. In Section~\ref{subsec-consistency}, using this pipeline, we identify a distinctive hallucination pattern: When objects mentioned in the image caption are replaced with discriminative statements and re-queried through the LVLM, it produces more accurate responses. Section~\ref{subsec-dependencies} examines how the absence of image input influences token generation. 

\subsection{Position-dependent hallucination}\label{subsec-patterns}
Through a simple visualization of caption generation, we observed that inaccurate descriptions tend to appear more frequently in later positions of the generated sequence. To quantify this observation, we utilized \textbf{M-HalDetect}~\cite{gunjal2024detecting}, a human-annotated dataset containing 16,000 image captions generated by \textbf{InstructBLIP}~\cite{dai2023instructblip}. This dataset classifies descriptions into four categories:
\begin{itemize}[label=\textbullet]
    \item \textbf{Accurate}: Descriptions that correctly reflect objects and relationships present in the image.
    \item \textbf{Inaccurate}: Descriptions containing non-existent objects or incorrect interpretations.
    \item \textbf{Analysis}: Subjective interpretations beyond direct visual evidence.
    \item \textbf{Unsure}: Cases where confident judgment is not possible.
\end{itemize}

\begin{figure}
    \centering
    \includegraphics[width=0.8\textwidth]{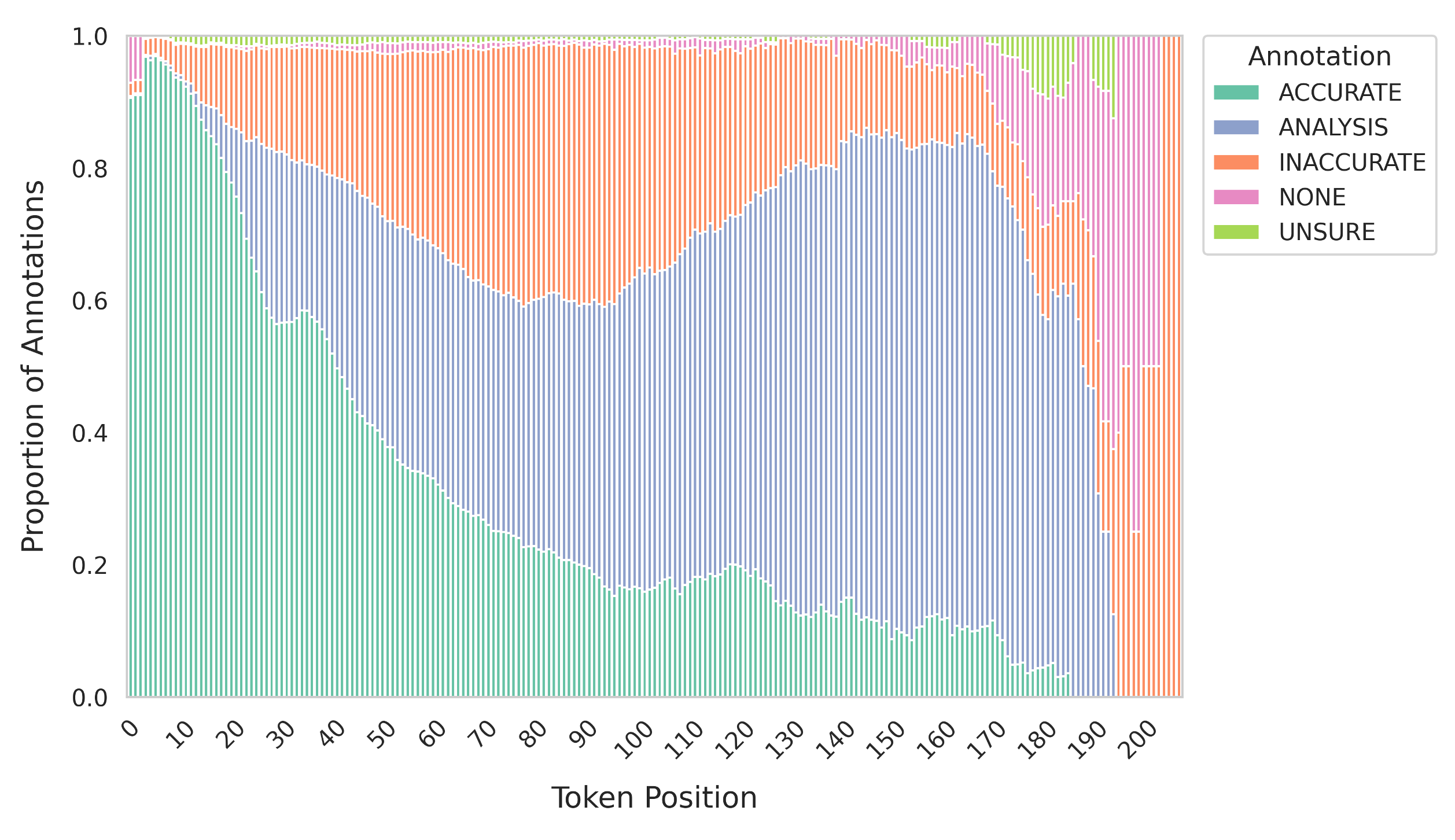} 
    \caption{The proportion of labels across token positions in the response}
    \label{fig:img1}
\end{figure}

Figure~\ref{fig:img1} shows that hallucinations occur predominantly in later portions of the responses. This pattern indicates two potential causes of hallucination:

First, hallucinations exhibit a snowball effect~\cite{zhong2024investigating}, where initial inaccuracies lead to progressively more unreliable content. The early hallucinations may serve as attention anchors, leading to an accumulation of related incorrect descriptions in subsequent generation steps.

Second, we observe that LVLMs tend to follow fixed response patterns learned during instruction fine-tuning. For example, LLaVA-1.5 consistently structures its responses with phrases \textquotedblleft This image features...\textquotedblright{} followed by \textquotedblleft In addition to...\textquotedblright{} in subsequent paragraphs. This template approach becomes problematic when handling images of varying complexity, as the model may prioritize completing these learned patterns over accurately describing the visual content, thereby introducing hallucinations.

\subsection{Object hallucination detection pipeline}\label{subsec-pipeline}
\begin{figure}
    \centering
    \includegraphics[width=0.8\textwidth]{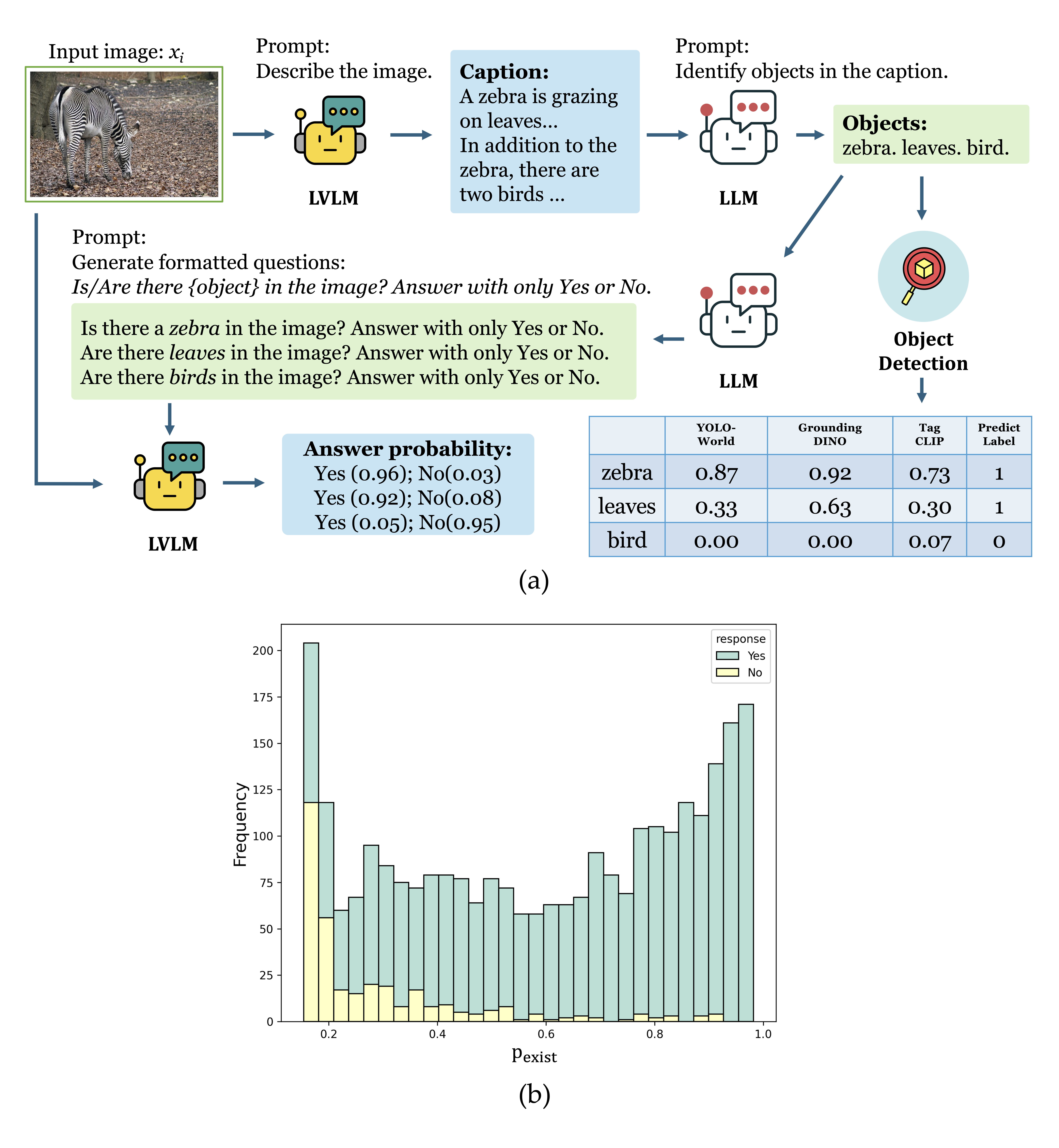} 
    \caption{(a) The workflow compares LLaVA's consistency in object existence judgments across two tasks; (b) The histogram displays the distribution of $p_{\text{exist}}$ values predicted by the Pipeline for Yes/No responses in discriminative tasks.}
    \label{fig:img2}
\end{figure}

To implement object hallucination detection without ground truth annotations, we reframe the task as a multi-object open-vocabulary detection problem. Our pipeline (illustrated in Figure~\ref{fig:img2}) first employs a Large Language Model (Llama-3.1-8B)\cite{dubey2024llama} to extract objects mentioned in captions using carefully designed prompts (detailed in Appendix~\ref{appendix-prompt}). The extracted objects then become detection targets for three open-vocabulary object detection (OVOD) tools to verify the presence of objects.

\textbf{YOLO-World~\cite{cheng2024yolo}:} The YOLO series traditionally excels in fast object detection but is constrained to a fixed set of predefined categories. YOLO-World overcomes this limitation by incorporating CLIP's text encoder and employing region-text contrastive learning on large-scale datasets. The key innovation is the Re-parameterizable Vision Language Path Aggregation Network, which fuses image and text features to enable zero-shot detection of novel object categories.

\textbf{Grounding DINO~\cite{liu2024grounding}:} Unlike models that rely on CLIP embeddings, Grounding DINO builds a transformer-based detector. The detector learns a unified vision language representation through extensive pre-training on image-text pairs with region annotations. This architecture enables precise localization and detection of multiple objects described by natural language queries.

\textbf{TagCLIP~\cite{lin2024tagclip}:} Standard CLIP models typically focus on dominant objects in images, limiting their ability to detect multiple objects. TagCLIP addresses this limitation through patch-level feature analysis, identifying less prominent objects by masking and re-scoring different image regions. This approach effectively supports multi-object detection in complex scenes.

These three tools provide complementary strengths for multi-object open-vocabulary detection. We combine their confidence scores using weights determined through logistic regression (detailed in Appendix~\ref{appendix-pipeline}) to achieve optimal object detection accuracy. The formulation is as follows:

\begin{equation}
\begin{split}
\text{logit}(p_{\text{exist}}) =\ & -1.7251 + 2.6723 \times \text{YOLO\_conf} \\
& + 1.6066 \times \text{DINO\_conf} + 2.2660 \times \text{TagCLIP\_conf}.
\end{split}
\end{equation}

Where $\text{logit}(p)$ represents the log-odds, which is defined as:

\begin{equation}
\text{logit}(p_{\text{exist}}) = \log \left( \frac{p_{\text{exist}}}{1 - p_{\text{exist}}} \right).
\end{equation}

To obtain the probability $p_{\text{exist}}$, we apply the sigmoid function:

\begin{equation}
p_{\text{exist}} = \frac{1}{1 + e^{-\text{logit}(p_{\text{exist}})}}.
\end{equation}

\subsection{Cross-task inconsistency}\label{subsec-consistency}
We discovered an intriguing pattern in the behavior of LVLMs: the same model can provide inconsistent predictions about the existence of objects in the same image between generative and discriminative tasks. To investigate this phenomenon, we performed an experiment following the workflow shown in Figure 2(a).

\begin{enumerate}
    \item Use LLaVA-1.5-7B to generate 500 image descriptions.
    \item Extract the mentioned objects using an LLM and generate formatted discriminative questions: \textquotedblleft Is/Are there \textless object\textgreater{} in the image? Answer with only Yes or No.\textquotedblright
    \item Input these questions back into LLaVA along with the original images to obtain Yes/No statistics.
\end{enumerate}

The results showed that in approximately 10\% of the cases, the response was \textit{No}, indicating conflicting judgments between generative and discriminative tasks. In Figure 2(b), we further compared the model's responses to the discriminative task with the \( p_{\text{exist}} \) calculated by the pipeline in Section~\ref{subsec-pipeline}. When inconsistencies occurred, the responses from the discriminative task typically aligned more closely with the predictions of the pipeline, as evidenced by the lower \( p_{\text{exist}} \) distribution for objects that were answered \textit{No} in discriminative tasks. This suggests two key insights:

\begin{itemize}
    \item Benchmarks that rely on discriminative tasks to evaluate model hallucination may underestimate the model's tendency to hallucinate in generative tasks.
    \item This cross-task inconsistency suggests that some hallucinations do not stem from misunderstanding image information, but rather from language priors dominating the text generation phase.
\end{itemize}

In particular, LLaVA-1.5 shows better performance in discriminative tasks, which can be attributed to its fine-tuning process using formatted prompts~\cite{liu2024improved}. This experiment shows that LVLMs are more prone to hallucination in open-ended generation tasks.

\subsection{Dependencies on visual input}\label{subsec-dependencies}
Previous studies have indicated that LVLMs can successfully answer VQA tasks without image inputs, suggesting that language priors significantly influence model behavior~\cite{chen2024we}. To measure the influence of visual information on token generation, we compared the token probability distributions between image-present and image-absent conditions at each generation step:

\begin{figure}
    \centering
    \includegraphics[width=0.8\textwidth]{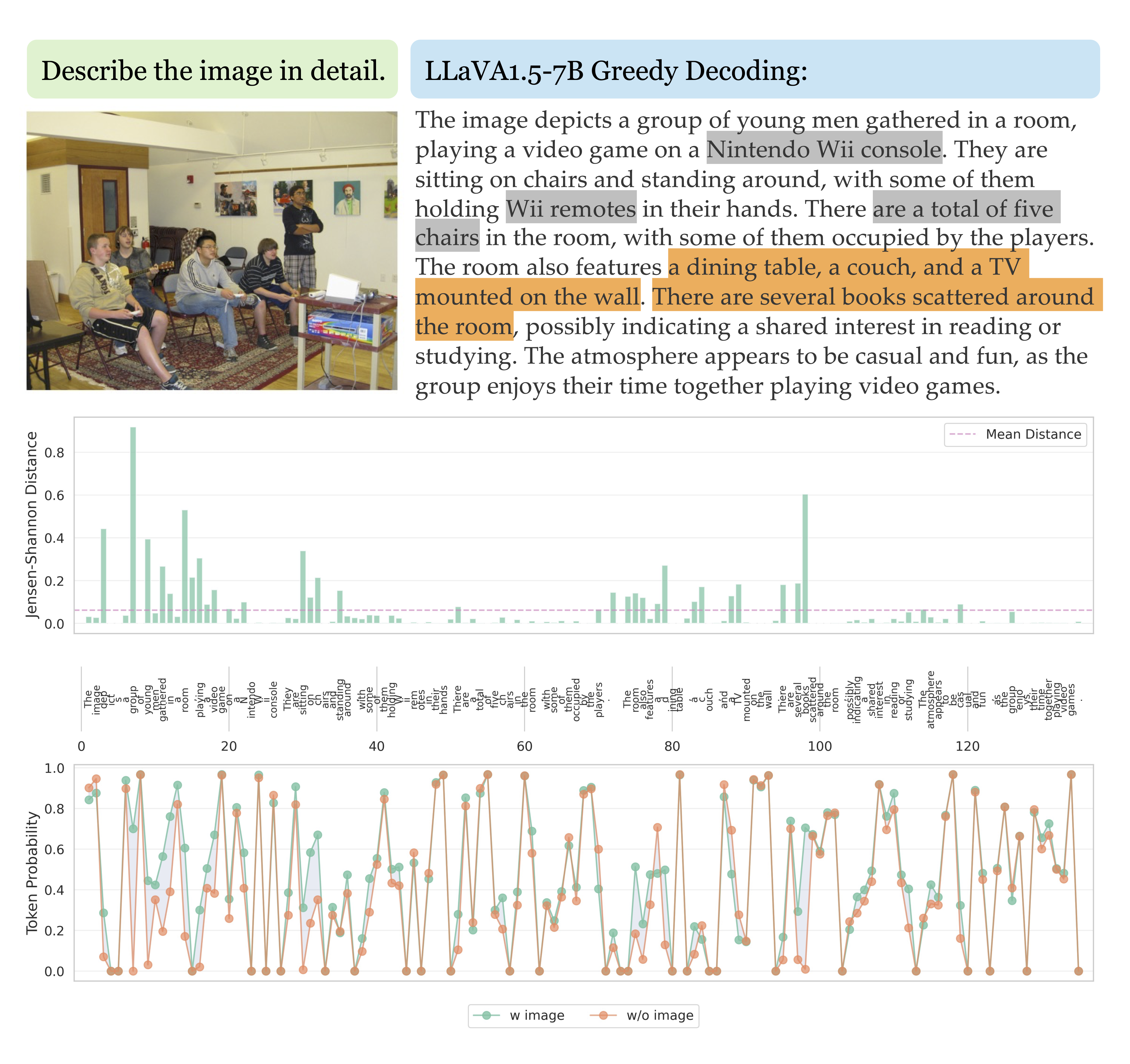} 
    \caption{Comparative visualization of token probabilities and distributions with and without image input. The caption annotations are manually labeled: gray indicates UNSURE, and orange highlights INACCURATE content.}
    \label{fig:img3}
\end{figure}

\vspace{\baselineskip}
\begin{enumerate}
    \item Generate an image caption with greedy decoding, selecting the highest probability token at each step for the image in Figure~\ref{fig:img3}.
    \item Use the forward function to generate the same output sequence without image input, and record the token probability distributions at each generation step.
    \item Analyze the token distribution divergence between with-image and without-image conditions using Jensen-Shannon Divergence (JSD). We scaled the JSD values to [0,1], where 0 represents identical distributions and 1 indicates maximum divergence.
\end{enumerate}

The middle graph of Figure~\ref{fig:img3} shows JSD values between token distributions, revealing that a substantial portion of tokens have low divergence values. The bottom graph compares selected token probabilities, demonstrating highly similar trajectories for many tokens regardless of visual input.

Our analysis reveals that LLaVA's generation process relies heavily on language priors once the initial context is established. Within individual sentences, while initial tokens show stronger dependence on visual features, the subsequent generation appears to be guided primarily by previous text. This finding provides indication of how language priors can override visual features during the generation process.

\section{Proposed method}\label{sec-proposedmethod}
In Section~\ref{sec-analyzingpatterns}, we observed that LVLMs often generate hallucinations in the later parts of the output (Section~\ref{subsec-patterns}) and that their token distributions can be quite similar with or without the actual image (Section~\ref{subsec-dependencies}). These findings suggest that, as the generation proceeds, the model relies increasingly on language priors rather than visual content.

To address this issue, we propose a token-level hallucination classifier that uses the model's internal hidden states—both with and without the image input—to detect whether each token is accurately grounded or not. During inference, we integrate this classifier into our decoding strategy to identify and remove potentially hallucinated tokens, employing top-$K$ sampling to generate multiple candidate sequences.

In the following sections, we first describe how we obtain the final hidden states for each token (Section~\ref{subsec-preliminary}). We then outline our data annotation procedure using the object hallucination detection pipeline (Section~\ref{subsec-annotation}) and detail the training of our token-level classifier (Section~\ref{subsec-classifier}). Finally, Section 5.4 explains how we incorporate this classifier into the generation process to reduce hallucinations in the output.

\subsection{Preliminaries}\label{subsec-preliminary}

\noindent
\textbf{Model overview.}  
Let $M_{\theta}$ denote a Vision Language Model parameterized by $\theta$. Given an image $v$ and a textual query $x$ as input, the model generates a textual output $y$ in an auto-regressive manner. At each decoding step $t$, the model obtains a final-layer hidden state $\mathbf{h}_t \in \mathbb{R}^d$, where $d$ (e.g., $d = 4096$ in a 7B model) denotes the dimensionality of the representation. The language head then transforms $\mathbf{h}_t$ into unnormalized logits $\mathbf{z}_t \in \mathbb{R}^{|V|}$:
\[
\mathbf{z}_t = \mathbf{W}_{\text{head}} \mathbf{h}_t + \mathbf{b}_{\text{head}},
\]
from which the next-token distribution is computed via softmax function:
\[
p_{\theta}(y_t | v, x, y_{<t}) = \operatorname{softmax}(\mathbf{z}_t).
\]
Thus, the token $y_t \in V$ is sampled from the vocabulary set $V$ according to this probability distribution.

\noindent
\textbf{Notation.}
To analyze the influence of the image $v$ on the model's behavior, we distinguish between the hidden states obtained when the image is provided and when it is not. We denote:
\begin{itemize}
    \item $X_1(t) = \mathbf{h}^{(v)}_t$: the final-layer hidden state at step $t$ with image $v$,
    \item $X_2(t) = \mathbf{h}^{(\neg v)}_t$: the final-layer hidden state at step $t$ without image $v$.
\end{itemize}

\noindent
\textbf{Measuring image dependence.}
To assess the model's dependence on visual input, we compare the difference between the hidden states in the two cases:
\[
\Delta \mathbf{h}(t) = \mathbf{h}^{(v)}_t - \mathbf{h}^{(\neg v)}_t = X_1(t) - X_2(t).
\]
A large norm $\Delta \mathbf{h}(t)$ indicates that the model heavily relies on the visual features from the image $v$ when generating the token at step $t$. Conversely, a small norm implies that the model primarily relies on its language priors.

\subsection{Caption annotation}\label{subsec-annotation}
To train our token-level classifier, we require an automated method to provide token-level annotations for captions, specifically a classifier that determines whether each token is part of hallucinated content. To accurately capture the model's true internal states during caption generation, we generate captions using \texttt{LLaVA-1.5} itself instead of relying on manually labeled captions from other models.

We generate captions for 2,000 images (distinct from those used in our evaluation in Section~\ref{sec-experiments}) using greedy decoding. We leverage the Object Hallucination Detection Pipeline described in Section~\ref{subsec-pipeline} to compute a confidence score \( p_{\text{exist}} \) for each object mentioned in the captions. Based on these scores, we classify objects as accurate or inaccurate. We then apply the following rules to annotate each token in the captions:
\begin{enumerate}[label=\arabic*.]
    \item Initially label all tokens as \textbf{ACCURATE}.
    \item For each inaccurate object, mark all tokens in its containing sentence (separated by periods) as \textbf{INACCURATE}.
    \item For each accurate object, if its phrase (separated by commas or periods) was previously marked as \textbf{INACCURATE}, reset those tokens to \textbf{ACCURATE}.
\end{enumerate}

This annotation process, while heuristic, allows us to efficiently label large amounts of training data at the token level.

\subsection{Token-level hallucination classifier}\label{subsec-classifier}
\begin{figure}[t]
    \centering
    \includegraphics[width=0.8\linewidth]{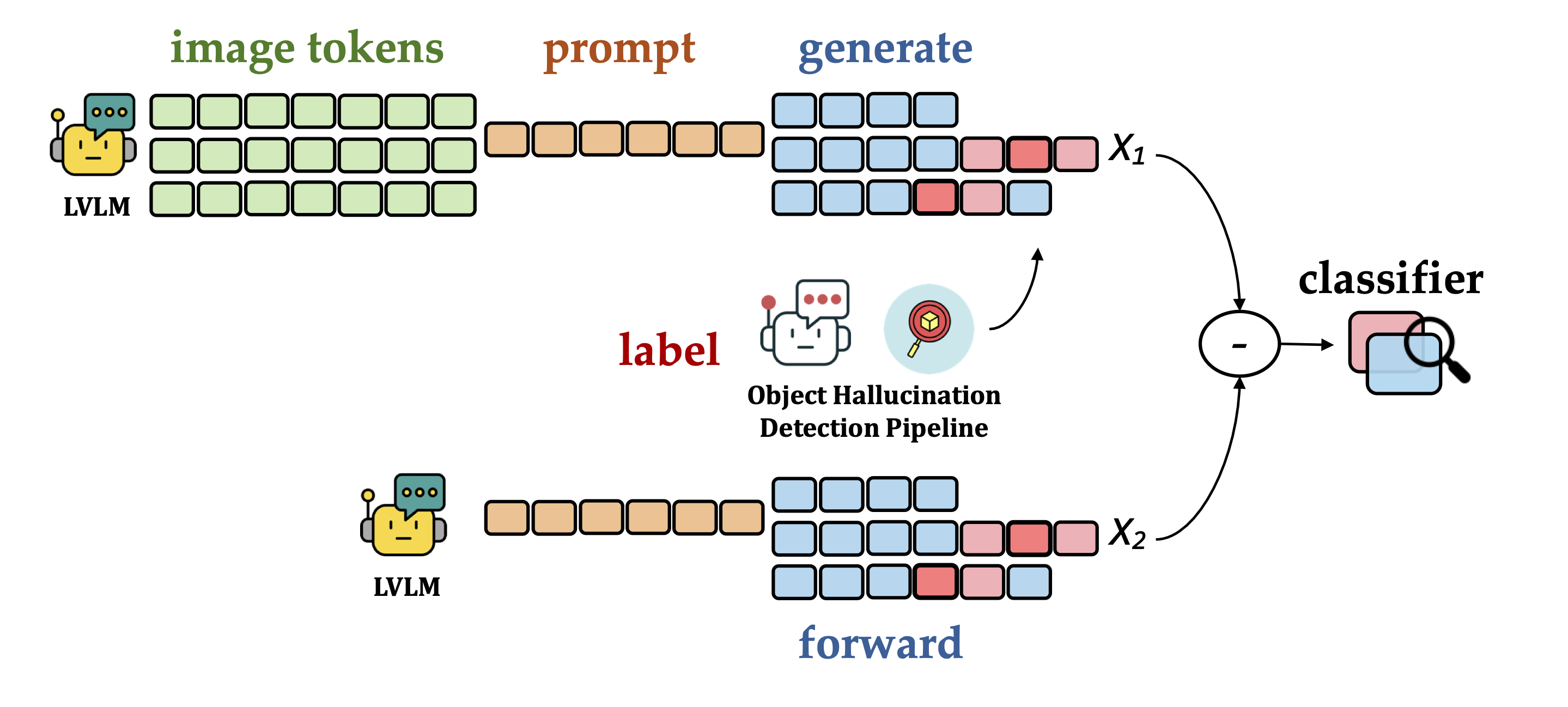} 
    \caption{Training process for token-level hallucination classifier.}
    \label{fig:img4}
\end{figure}

Using the annotated captions from Section~\ref{subsec-annotation}, we train a simple Multi-Layer Perceptron (MLP) classifier to predict whether each token is hallucinated based on the model's hidden states. Figure~\ref{fig:img4} illustrates the classifier training workflow. For each token position \( t \), we collect:
\begin{align*}
X_1(t) & : \text{The hidden state with image input}, \\
X_2(t) & : \text{The hidden state without image input}, \\
y(t)   & : \text{The binary label from our annotation process}.
\end{align*}

We choose hidden states in the final layer instead of token probability distributions because they offer two advantages: a more compact representation (dimension \( d \) vs. vocabulary size \( |V| \)) and richer encoded information that includes positional context.

The classifier consists of a 3-layer MLP with hidden dimensions \((256, 128, 64)\), where each hidden layer includes batch normalization, ReLU activation, and dropout \((0.5)\). The final layer outputs a single logit, which is passed through a sigmoid function to obtain the hallucination probability. We experiment with three different input features: 
\begin{itemize}
    \item Only \( X_1 \),
    \item Only \( X_2 \),
    \item The difference \( X_1 - X_2 \).
\end{itemize}

Table~\ref{tab:feature-comparison} shows that using only \( X_2 \) achieves an above-baseline performance, probably because the model learns position-dependent hallucination patterns. Using only \( X_1 \) performs better than \( X_2 \), while the difference \( X_1 - X_2 \) yields the best classification accuracy. The results support our hypothesis: the difference between these hidden states emphasizes how the model's internal representation is altered by image features, thereby providing a stronger signal to identify when hallucinations occur.

\begin{table}[t]
\centering
\caption{Performance comparison of different training feature. The "Majority" baseline selects the majority class for all predictions. 
Values in bold represent the best performance across the input features.}
\label{tab:feature-comparison}
\begin{tabular}{lcccccc}
\toprule
\multirow{2}{*}{Input Features} & \multicolumn{3}{c}{ACCURATE Class} & \multicolumn{3}{c}{INACCURATE Class} \\
\cmidrule(lr){2-4} \cmidrule(lr){5-7}
 & Precision & Recall & F1 & Precision & Recall & F1 \\
\midrule
Majority & 0.7306 & 1.0000 & 0.8443 & 0.0000 & 0.0000 & 0.0000 \\
$X_{2}$  & 0.8607 & 0.8883 & 0.8743 & 0.6682 & 0.6100 & 0.6378 \\
$X_{1}$  & 0.8693 & 0.8942 & 0.8816 & 0.6889 & 0.6353 & 0.6610 \\
$X_{1}-X_{2}$ & \textbf{0.8837} & \textbf{0.9274} & \textbf{0.9050} & \textbf{0.7725} & \textbf{0.6689} & \textbf{0.7169} \\
\bottomrule
\end{tabular}
\end{table}

Our classifier achieves promising accuracy in predicting token-level hallucinations, demonstrating its potential to identify accurate content during the generation process. Detailed model architecture and training parameters are provided in Appendix~\ref{appendix-trainingclassifier}.

\subsection{Decoding strategy}\label{subsec-decodingstrategy}

Our proposed decoding strategy for generating accurate image captions consists of three main components: Top-$K$ First Token Sampling (Algorithm 1) for generating diverse candidate sentences, Compute Sentence ACCURate (Algorithm 2) for evaluating the accuracy of each candidate, and the overall Sentence-Level Decoding (Algorithm 3) which iteratively combines these methods to produce the final caption.

\noindent
\textbf{Top-$K$ first token sampling (Algorithm 1)} generates a diverse set of $K$ candidate sentences. The algorithm first conducts a single forward pass through the language vision language model (LVLM) $M_{\theta}$, conditioned on the given image $v$ and text prefix, to obtain the $K$ most probable next tokens. For each sampled first token $y_1^{(i)}$, the algorithm then performs greedy decoding until a stopping condition (either a period or an end-of-sequence token <EOS>) is met, resulting in a complete candidate sentence $c_i$. 

\begin{algorithm}
\caption{Top-$K$ first token sampling}
\label{alg:topK}
\KwIn{ 
LVLM $M_{\theta}$, Image $v$, Text prefix \textit{prefix}, Sampling size $K$ 
}
\KwOut{Candidate sentence list $\{c_1, \dots, c_K\}$}
\BlankLine
Initialize $\{c_1, \dots, c_K\} \gets \{\}$\;

\tcp{Step (a): Sample top-$K$ next tokens in one forward pass}
Sample top $K$ next tokens $\{y_1^{(1)}, \dots, y_1^{(K)}\}$ from $M_{\theta}$ conditioned on \textit{prefix} and $v$\;

\For{$i \gets 1$ \KwTo $K$}{
    \tcp{Step (b): Greedy decoding until stopping condition}
    Starting from $y_{1}^{(i)}$, greedily decode until a period (.) or \texttt{<EOS>} is encountered, producing $c_i$\;
}

\Return $\{c_1, \dots, c_K\}$\;
\end{algorithm}

\noindent
\textbf{Compute sentence ACCURate (Algorithm 2)} assesses the accuracy of a candidate sentence using the token-level hallucination classifier introduced in Section~\ref{subsec-classifier}. The algorithm first concatenates the text prefix and the candidate sentence to form a complete input sequence. It then performs a forward pass through the LVLM to extract the last hidden layer representations for each token, with and without the image input. 

\begin{figure}[t]
    \centering
    \includegraphics[width=0.6\linewidth]{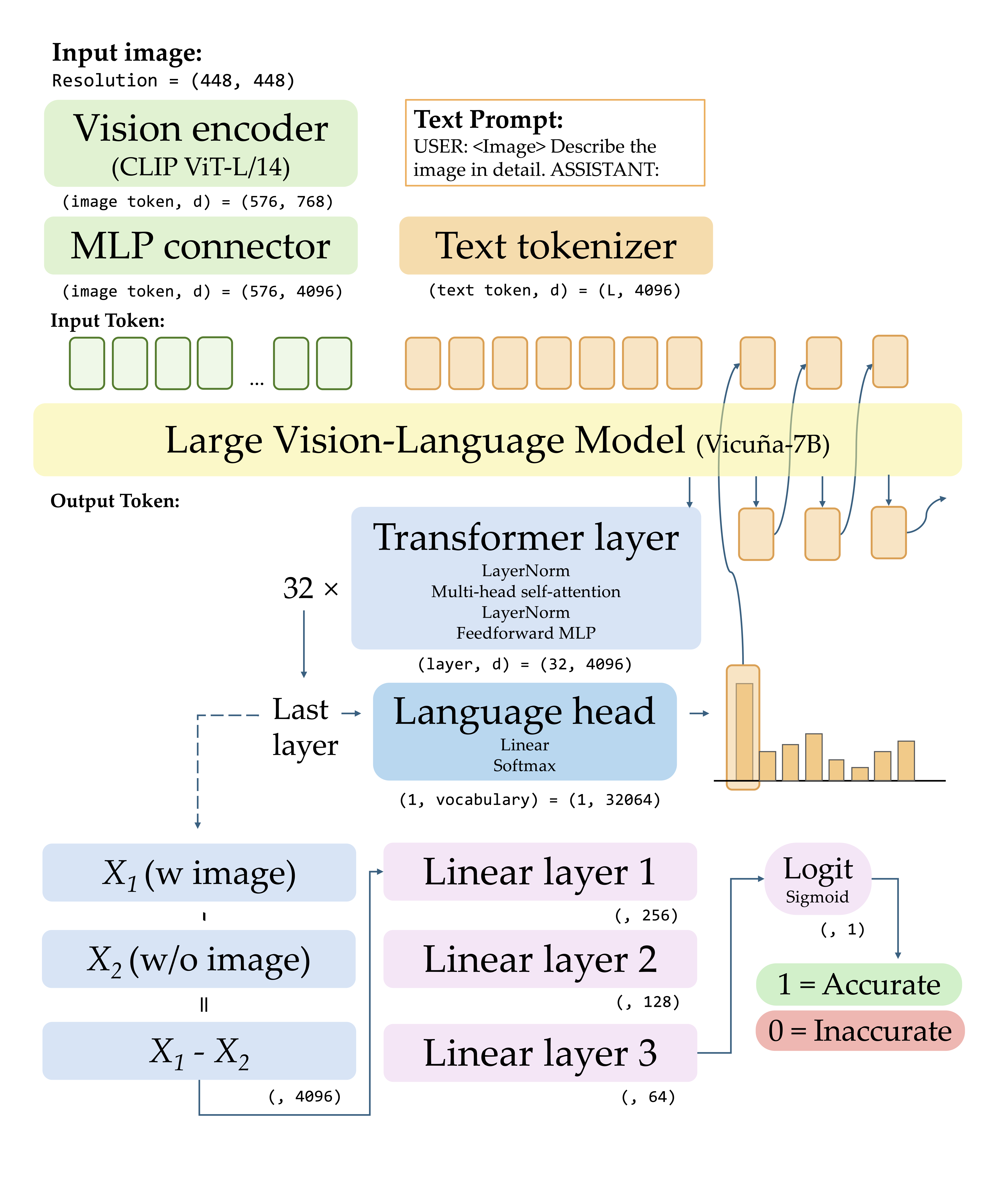} 
    \caption{Token-level classification at inference time.}
    \label{fig:img5}
\end{figure}

For each token $y_t$ in the candidate sentence, the classifier predicts whether it is classified as \texttt{Accurate} or \texttt{Inaccurate}. Figure~\ref{fig:img5} illustrates the classification process at inference time. And the sentence-level accuracy $ACCURate$ is computed by calculating the proportion of tokens classified as \texttt{Accurate}, providing an overall accuracy measure for the candidate sentence.

\begin{algorithm}[t]
\caption{Compute sentence ACCURate}
\label{alg:computeAcc}
\KwIn{
LVLM $M_{\theta}$, Image $v$, Text prefix \text{prefix}, Candidate sentence \textit{c}, Token classifier \text{Classifier}($\cdot$)
}
\KwOut{$\text{ACCURate}(c)$: accuracy of the sentence}
\BlankLine
\tcp{Step (a): Concatenate prefix and candidate sentence}
Concatenate \textit{prefix} and \textit{c} into input sequence $y_{<t}$\;

\tcp{Step (b): Compute last hidden layer representations with and without image $v$}
Use $M_{\theta}$ forward pass to compute:
\[
X_{1}(t) = \mathbf{h}^{(v)}_t,\quad X_{2}(t) = \mathbf{h}^{(\neg v)}_t \quad \text{for each token } t \text{ in } c.
\]

\tcp{Step (c): Classify each token as Accurate or Inaccurate}
For each token $y_t$ in \textit{c}, compute:
\[
\ell_t = \text{Classifier}(X_{1}(t), X_{2}(t)) \quad \in \{\texttt{Accurate}, \texttt{Inaccurate}\}.
\]

\tcp{Step (d): Calculate the proportion of Accurate tokens}
\[
\text{ACCURate}(c) = \frac{\sum_{t=1}^{|\textit{c}|} \mathbf{1}[\ell_t = \texttt{Accurate}]}{|\textit{c}|}
\]

\Return $\text{ACCURate}(c)$\;
\end{algorithm}

\noindent
\textbf{Sentence-level decoding (Algorithm 3)} generates the final caption by iteratively combining the methods from Algorithms 1 and 2. It first uses Top-$K$ First Token Sampling (Algorithm 1) to generate a set of candidate sentences. It then evaluates the accuracy of each remaining candidate using the Compute Sentence ACCURate method (Algorithm 2) and selects the one with the highest score. This process repeats, appending the selected candidates to form the caption, until any candidate ends with an end-of-sequence token. Finally, the algorithm filters out any sentences with an ACCURate below the specified threshold $t$ and concatenates the rest to produce the final, accurately generated caption while removing potentially hallucinated content.

\begin{algorithm}
\caption{Sentence-level decoding}
\label{alg:decoding_modified}
\KwIn{
LVLM $M_{\theta}$, Image $v$, Token classifier \textit{Classifier}, Threshold $t$, Sampling size $K$
}
\KwOut{Final caption}
\BlankLine
Initialize $sents \gets [\,]$ \tcp{List of generated sentences}  
Initialize $accu \gets [\,]$ \tcp{List of ACCURate scores}  
Initialize $prefix \gets \varnothing$ \tcp{Empty text prefix}  

\While{\textbf{True}}{
    \tcp{Step (a): Generate K candidate sentences}
    $C \gets \texttt{TopKFirstTokenSampling}(M_{\theta}, v, prefix, K)$\;

    \tcp{Step (b): Compute ACCURate for each candidate sentence}
    Initialize $C_{accurate} \gets [\,]$ \;
    
    \ForEach{$c \in C$}{
        $\alpha \gets \texttt{ComputeACCRate}(M_{\theta}, v, prefix, c, Classifier)$\;
        Append $(c, \alpha)$ to $C_{accurate}$\;
    }
    
    \tcp{Step (c): Select candidate sentence with highest ACCURate}
    $c^* \gets \arg\max(\alpha)$ for $(c, \alpha) \in C_{accurate}$\;  
    $\alpha^* \gets \max(\alpha)$ for $(c, \alpha) \in C_{accurate}$\;  
    Append $c^*$ to $sents$ and $\alpha^*$ to $accu$\;  
    Update $prefix \gets prefix + c^*$\;  
    
    \If{\(\exists\, c \in C \ \text{such that}\ c \ \text{contains}\ \texttt{<EOS>}\)}{
        \textbf{break} \tcp{Terminate once any candidate includes <EOS>}
    }
}

\tcp{Step (d): Filter sentences based on threshold $t$}
Initialize $FinalSents \gets [\,]$\;  
\For{$i \gets 1$ \KwTo $|sents|$}{
    \If{$accu[i] \ge t$}{
        Append $sents[i]$ to $FinalSents$\;
    }
}

\tcp{Step (e): Concatenate filtered sentences into the final caption}
$caption \gets \texttt{join}(FinalSents)$\;
\Return $caption$\;
\end{algorithm}

\section{Experiments}\label{sec-experiments}

\subsection{Datasets}\label{subsec-datasets}

We evaluate our proposed decoding approach by comparing it against three other methods:  
greedy decoding, OPERA~\cite{huang2024opera}, and VCD~\cite{leng2024mitigating}.  
Captions are generated using the LLaVA-1.5-7B model combined with each of these decoding techniques  
on a subset of 500 randomly selected images from the COCO 2014 validation set~\cite{lin2014microsoft}.

It is important to note that while OPERA and VCD are fully training-free methods that operate  
without additional tools or fine-tuning of the LVLM, our approach incorporates OVOD tools in  
the annotation pipeline and includes a separately trained classifier. After the initial  
classifier training, our pipeline functions as a fully automated system that can be efficiently  
deployed without requiring any model fine-tuning.

\subsection{Evaluation metrics}\label{subsec-eval}

The CHAIR (Captions with Hallucinated and Accurate Image Reporting) metric~\cite{rohrbach2018object} is a widely used framework for evaluating object hallucinations in image captions by comparing generated captions with ground-truth annotations. It operates by first taking the segmentation annotations provided by the COCO dataset (described in Section 6.1) as the ground-truth objects. The process then extracts objects mentioned in the captions through semantic parsing and maps them to COCO object categories. Subsequently, the CHAIR metric computes the following parameters:

\textbf{\(\mathbf{CHAIR_s}\):} Measures the percentage of captions containing hallucination across a dataset. A caption is considered to contain hallucination when it mentions any object that is not present in the corresponding image. It is calculated as:
\[
\mathbf{CHAIR_s} = \frac{\lvert \text{hallucinated captions} \rvert}{\lvert \text{all captions} \rvert}.
\]

\textbf{\(\mathbf{CHAIR_i}\):} Measures the proportion of hallucinated objects within individual captions by calculating the ratio of hallucinated objects to the total number of objects mentioned in each caption. It is calculated as:
\[
\mathbf{CHAIR_i} = \frac{\lvert \text{hallucinated objects} \rvert}{\lvert \text{all mentioned objects} \rvert}.
\]

To ensure a comprehensive evaluation, the framework includes two additional supporting metrics:

\textbf{Recall:} Evaluates how many objects from the ground truth annotations are mentioned in the generated caption, ensuring that hallucination reduction does not come at the cost of missing image content.

\textbf{Length:} Tracks the word count of the captions as a basic reference for descriptive detail.

\subsection{Results}\label{subsec-result}

The experimental results in Table~\ref{tab:chair} show that the proposed method significantly reduces $\mathbf{CHAIR}_{s}$ and $\mathbf{CHAIR}_{i}$ compared to baselines. The threshold parameter $t$ offers control over the hallucination-accuracy trade-off: Higher thresholds reduce hallucinations but may exclude some accurate descriptions. Caption examples generated by these decoding methods are provided in Appendix~\ref{appendix-captionsample}.

\begin{table}[t]
    \centering
    \caption{CHAIR metric comparison of different decoding methods. \textit{K} denotes the top-$K$ sample size, and \textit{t} represents the threshold parameter controlling the sentence hallucination rate.}
    \label{tab:chair}
    \begin{tabular}{lcccc}
        \toprule
        LLaVA-1.5-7B & CHAIR$_s$ $\downarrow$ & CHAIR$_i$ $\downarrow$ & Recall $\uparrow$ & Len \\
        \midrule
        \midrule
        Greedy       & 47.2 & 13.6 & 79.5 & 88.9 \\
        OPERA        & 44.6 & 12.7 & 79.2 & 92.7 \\
        VCD          & 53.8 & 15.0 & 77.0 & 98.5 \\
        $K = 1, t = 0.5$ & 34.0 & 9.9 & 73.8 & 76.6 \\
        $K = 1, t = 0.6$ & 28.8 & 8.6 & 70.6 & 70.0 \\
        $K = 3, t = 0.6$ & 27.0 & 7.7 & 71.2 & 71.0 \\
        $K = 1, t = 0.7$ & 22.2 & 7.1 & 66.3 & 62.6 \\
        $K = 3, t = 0.7$ & 22.2 & 6.4 & 68.2 & 65.2 \\
        $K = 1, t = 0.8$ & 16.2 & 5.3 & 61.1 & 54.5 \\
        $K = 3, t = 0.8$ & 17.8 & 5.3 & 63.7 & 58.5 \\
        \bottomrule
    \end{tabular}
\end{table}

On the other hand, top-$K$ sampling showed limited performance improvement, this may result from insufficient diversity in the sampled sentences or limitations in the classifier's token prediction accuracy. Our findings, supported by Table~\ref{tab:feature-comparison}, suggest that hallucinations often occur when the model generates text without conditioning on the image input. This pattern can be detected by analyzing differences in the model's hidden states when prompted with and without images.

\section{Conclusions}\label{sec-conclusion}
Our experimental results reveal that Large Vision Language Models (LVLMs) are more prone to hallucinations when generating open-ended image captions compared to answering discriminative questions. Through analyzing the token probability distributions at each generation step, we found a pronounced similarity between distributions with and without image input, except for the beginning part of captions and the initial tokens of some subsequent sentence. This suggests that the previously generated sequence significantly conditions the subsequent token distributions, making the model’s outputs heavily dependent on language context.

This observation illuminates the mechanism of hallucination in LVLMs. Our experiments suggest that many hallucinations emerge not from incorrect visual feature encoding or insufficient vision-language alignment, but rather due to language priors gradually overriding visual information during caption generation. The process shifts from image-guided text generation to purely language-driven next token prediction.

Building on these insights, we developed a top-down approach that uses the similarity between hidden layer vectors (with and without image input) as an indicator of potential hallucinations. Evaluation results show that this approach effectively reduces hallucinations in LVLM-generated captions, leading to more accurate and image-aligned outputs. Nevertheless, several limitations of this study should be acknowledged. 

\subsection{Limitations}\label{subsec-limitation}
First, this approach may mistakenly penalize cases where the description of visual content naturally aligns with statistical patterns learned from the training data. Second, our current implementation relies on basic vector subtraction for comparing hidden layer representations, without exploring more sophisticated distance metrics, such as Euclidean distance or cosine similarity.

It is important to note that the hidden layer vectors we use are highly model-specific and task-dependent. Consequently, training a classifier on captions generated by another model (e.g., InstructBLIP) — even if they are better annotated — may not yield improved performance. Furthermore, our classifier appears to learn correlations between token positions and hallucination probability specific to the image captioning task, limiting its applicability to other types of output. This in fact motivated our design of an automated hallucination annotation pipeline.

Finally, our token-level classifier was initially designed with the purpose of assessing sentence-level accuracy, which leads to some limitations. By labeling all tokens in a sentence as \textit{INACCURATE} when hallucinated objects are detected, we not only focused narrowly on object hallucinations but also reduced the classifier's ability to make fine-grained judgments about hallucinated content. We believe that developing better hallucination detection pipelines could result in a classifier with higher accuracy and enhanced overall performance.

\subsection{Discussion}\label{subsec-discussion}
Having examined the technical aspects of our approach and its limitations, we now turn to broader implications for LVLM development and training. Despite substantial instruction tuning on image captioning tasks, LVLMs like LLaVA continue to produce a significant amount of hallucinated content. We hypothesize that this might be related to the response template hypothesis discussed in Section~\ref{subsec-patterns}, where synthetic training data (such as GPT-4 generated captions) may contain fixed patterns that amplify the role of language priors in captioning. In our implementation of decoding methods, we found that incorporating top-$K$ first-token sampling did not introduce sufficient diversity. The template effect extends beyond just the highest-probability initial token, as alternative tokens also followed fixed order.

Image captioning serves as a fundamental task in evaluating LVLM capabilities and has received extensive attention. However, the mechanisms behind hallucinations in image captioning may differ slightly from those in broader VQA or reasoning tasks. This suggests the need for generalized end-to-end solutions to address hallucinations in LVLMs. We leave this challenge to future research and look forward to new hypotheses and solutions from the research community.

\bibliographystyle{abbrv}
\bibliography{references}

\begin{thebibliography}{10}

\bibitem{achiam2023gpt}
J.~Achiam, S.~Adler, S.~Agarwal, L.~Ahmad, I.~Akkaya, F.~L. Aleman, D.~Almeida, J.~Altenschmidt, S.~Altman, S.~Anadkat, et~al.
\newblock {GPT-4} technical report.
\newblock In {\em Proceedings of the International Conference on Machine Learning (ICML)}, 2023.

\bibitem{chen2024we}
L.~Chen, J.~Li, X.~Dong, P.~Zhang, Y.~Zang, Z.~Chen, H.~Duan, J.~Wang, Y.~Qiao, D.~Lin, and F.~Zhao.
\newblock Are we on the right way for evaluating large vision-language models?
\newblock In {\em Proceedings of the Thirty-Eighth Annual Conference on Neural Information Processing Systems (NeurIPS)}, 2024.

\bibitem{chen2024multi}
X.~Chen, Z.~Ma, X.~Zhang, S.~Xu, S.~Qian, J.~Yang, D.~Fouhey, and J.~Chai.
\newblock Multi-object hallucination in vision language models.
\newblock In {\em Proceedings of the Thirty-Eighth Annual Conference on Neural Information Processing Systems (NeurIPS)}, 2024.

\bibitem{cheng2024yolo}
T.~Cheng, L.~Song, Y.~Ge, W.~Liu, X.~Wang, and Y.~Shan.
\newblock {YOLO-World}: Real-time open-vocabulary object detection.
\newblock In {\em 2024 IEEE/CVF Conference on Computer Vision and Pattern Recognition (CVPR)}, pages 16901--16911, 2024.

\bibitem{vicuna2023}
W.-L. Chiang, Z.~Li, Z.~Lin, Y.~Sheng, Z.~Wu, H.~Zhang, L.~Zheng, S.~Zhuang, Y.~Zhuang, J.~E. Gonzalez, I.~Stoica, and E.~P. Xing.
\newblock Vicuna: An open-source chatbot impressing {GPT-4} with 90\%* {ChatGPT} quality, March 2023.

\bibitem{dai2023instructblip}
W.~Dai, J.~Li, D.~Li, A.~M.~H. Tiong, J.~Zhao, W.~Wang, B.~Li, P.~Fung, and S.~Hoi.
\newblock {InstructBLIP}: Towards general-purpose vision-language models with instruction tuning.
\newblock In {\em Proceedings of the 37th International Conference on Neural Information Processing Systems (NeurIPS)}, 2023.

\bibitem{dubey2024llama}
A.~Dubey, A.~Jauhri, A.~Pandey, A.~Kadian, A.~Al-Dahle, A.~Letman, A.~Mathur, A.~Schelten, A.~Yang, A.~Fan, et~al.
\newblock The {Llama} 3 herd of models.
\newblock {\em arXiv preprint arXiv:2407.21783}, 2024.

\bibitem{gunjal2024detecting}
A.~Gunjal, J.~Yin, and E.~Bas.
\newblock Detecting and preventing hallucinations in large vision language models.
\newblock In {\em Proceedings of the Thirty-Eighth AAAI Conference on Artificial Intelligence}, pages 2023:1--2023:9, 2024.

\bibitem{liu2024survey}
L.~Huang, W.~Yu, W.~Ma, W.~Zhong, Z.~Feng, H.~Wang, Q.~Chen, W.~Peng, X.~Feng, B.~Qin, and T.~Liu.
\newblock A survey on hallucination in large language models: Principles, taxonomy, challenges, and open questions.
\newblock {\em ACM Transactions on Information Systems}, 43(2):42:1--42:55, 2025.

\bibitem{huang2024opera}
Q.~Huang, X.~Dong, P.~Zhang, B.~Wang, C.~He, J.~Wang, D.~Lin, W.~Zhang, and N.~Yu.
\newblock {OPERA}: Alleviating hallucination in multi-modal large language models via over-trust penalty and retrospection-allocation.
\newblock In {\em 2024 IEEE/CVF Conference on Computer Vision and Pattern Recognition (CVPR)}, pages 13418--13427, 2024.

\bibitem{leng2024mitigating}
S.~Leng, H.~Zhang, G.~Chen, X.~Li, S.~Lu, C.~Miao, and L.~Bing.
\newblock Mitigating object hallucinations in large vision-language models through visual contrastive decoding.
\newblock In {\em 2024 IEEE/CVF Conference on Computer Vision and Pattern Recognition (CVPR)}, pages 13872--13882, 2024.

\bibitem{li2023blip}
J.~Li, D.~Li, S.~Savarese, and S.~Hoi.
\newblock {BLIP-2}: Bootstrapping language-image pre-training with frozen image encoders and large language models.
\newblock In {\em Proceedings of the 40th International Conference on Machine Learning (ICML)}, 2023.

\bibitem{li2023evaluating}
Y.~Li, Y.~Du, K.~Zhou, J.~Wang, X.~Zhao, and J.-R. Wen.
\newblock Evaluating object hallucination in large vision-language models.
\newblock In {\em Proceedings of the 2023 Conference on Empirical Methods in Natural Language Processing}, pages 292--305, 2023.

\bibitem{lin2014microsoft}
T.-Y. Lin, M.~Maire, S.~Belongie, J.~Hays, P.~Perona, D.~Ramanan, P.~Doll{\'a}r, and C.~L. Zitnick.
\newblock {Microsoft COCO}: Common objects in context.
\newblock In {\em Proceedings of the European Conference on Computer Vision (ECCV)}, pages 740--755, 2014.

\bibitem{lin2024tagclip}
Y.~Lin, M.~Chen, K.~Zhang, H.~Li, M.~Li, Z.~Yang, D.~Lv, B.~Lin, H.~Liu, and D.~Cai.
\newblock {TagCLIP}: A local-to-global framework to enhance open-vocabulary multi-label classification of clip without training.
\newblock In {\em Proceedings of the Thirty-Eighth AAAI Conference on Artificial Intelligence}, pages 391--399, 2024.

\bibitem{liu2024improved}
H.~Liu, C.~Li, Y.~Li, and Y.~J. Lee.
\newblock Improved baselines with visual instruction tuning.
\newblock In {\em Proceedings of the IEEE/CVF Conference on Computer Vision and Pattern Recognition (CVPR)}, pages 26296--26306, 2024.

\bibitem{liu2024visual}
H.~Liu, C.~Li, Q.~Wu, and Y.~J. Lee.
\newblock Visual instruction tuning.
\newblock In {\em Proceedings of the 37th Conference on Neural Information Processing Systems (NeurIPS)}, 2023.

\bibitem{liu2024grounding}
S.~Liu, Z.~Zeng, T.~Ren, F.~Li, H.~Zhang, J.~Yang, Q.~Jiang, C.~Li, J.~Yang, H.~Su, J.~Zhu, and L.~Zhang.
\newblock {Grounding DINO}: Marrying dino with grounded pre-training for open-set object detection.
\newblock In {\em Computer Vision – ECCV 2024: 18th European Conference, Milan, Italy, Proceedings, Part XLVII}, pages 38--55, 2024.

\bibitem{radford2021learning}
A.~Radford, J.~W. Kim, C.~Hallacy, A.~Ramesh, G.~Goh, S.~Agarwal, G.~Sastry, A.~Askell, P.~Mishkin, J.~Clark, G.~Krueger, and I.~Sutskever.
\newblock Learning transferable visual models from natural language supervision.
\newblock In {\em Proceedings of the 38th International Conference on Machine Learning (ICML)}, volume 139, pages 8748--8763, 2021.

\bibitem{rohrbach2018object}
A.~Rohrbach, L.~A. Hendricks, K.~Burns, T.~Darrell, and K.~Saenko.
\newblock Object hallucination in image captioning.
\newblock In {\em Proceedings of the 2018 Conference on Empirical Methods in Natural Language Processing}, pages 4035--4045, 2018.

\bibitem{team2023gemini}
G.~Team, R.~Anil, S.~Borgeaud, J.-B. Alayrac, J.~Yu, R.~Soricut, et~al.
\newblock Gemini: A family of highly capable multimodal models.
\newblock {\em arXiv e-prints}, 2023.

\bibitem{wang2023llm}
J.~Wang, Y.~Wang, G.~Xu, J.~Zhang, Y.~Gu, H.~Jia, M.~Yan, J.~Zhang, and J.~Sang.
\newblock An {LLM}-free multi-dimensional benchmark for mllms hallucination evaluation.
\newblock {\em arXiv preprint arXiv:2311.07397}, 2023.

\bibitem{yin2024woodpecker}
S.~Yin, C.~Fu, S.~Zhao, T.~Xu, H.~Wang, D.~Sui, Y.~Shen, K.~Li, X.~Sun, and E.~Chen.
\newblock Woodpecker: Hallucination correction for multimodal large language models.
\newblock {\em Science China Information Sciences}, 67(12):220105, 2024.

\bibitem{zhang2024visually}
Y.~Zhang, A.~Unell, X.~Wang, D.~Ghosh, Y.~Su, L.~Schmidt, and S.~Yeung-Levy.
\newblock Why are visually-grounded language models bad at image classification?
\newblock {\em arXiv preprint arXiv:2405.18415}, 2024.

\bibitem{zhao2023beyond}
Z.~Zhao, B.~Wang, L.~Ouyang, X.~Dong, J.~Wang, and C.~He.
\newblock Beyond hallucinations: Enhancing lvlms through hallucination-aware direct preference optimization.
\newblock {\em arXiv preprint arXiv:2311.16839}, 2023.

\bibitem{zhong2024investigating}
W.~Zhong, X.~Feng, L.~Zhao, Q.~Li, L.~Huang, Y.~Gu, W.~Ma, Y.~Xu, and B.~Qin.
\newblock Investigating and mitigating the multimodal hallucination snowballing in large vision-language models.
\newblock In {\em Proceedings of the 62nd Annual Meeting of the Association for Computational Linguistics (Volume 1: Long Papers)}, pages 11991--12011, 2024.

\bibitem{zhou2023analyzing}
Y.~Zhou, C.~Cui, J.~Yoon, L.~Zhang, Z.~Deng, C.~Finn, M.~Bansal, and H.~Yao.
\newblock Analyzing and mitigating object hallucination in large vision-language models.
\newblock In {\em Proceedings of the Twelfth International Conference on Learning Representations (ICLR)}, 2024.

\bibitem{zhu2023minigpt}
D.~Zhu, J.~Chen, X.~Shen, X.~Li, and M.~Elhoseiny.
\newblock {MiniGPT-4:} enhancing vision-language understanding with advanced large language models.
\newblock In {\em Proceedings of the Twelfth International Conference on Learning Representations (ICLR)}, 2024.

\end{thebibliography}

\clearpage
\appendix  
\section*{Appendix}

\begin{table}[t]
    \centering
    \caption{Details of the models used in the study.}
    \label{tab:model-data}
    \begin{tabular}{|p{4cm}|p{10cm}|} 
        \hline
        \textbf{Model} & \textbf{Link} \\
        \hline
        LLaVA-1.5-7b~\cite{liu2024improved} & \url{https://huggingface.co/liuhaotian/llava-v1.5-7b} \\
        Llama-3.1-8B~\cite{dubey2024llama} & \url{https://huggingface.co/meta-llama/Llama-3.1-8B} \\
        YOLO-World~\cite{cheng2024yolo} & \url{https://github.com/AILab-CVC/YOLO-World} \\
        Grounding DINO~\cite{liu2024grounding} & \url{https://huggingface.co/IDEA-Research/grounding-dino-base} \\
        TagCLIP~\cite{lin2024tagclip} & \url{https://github.com/linyq2117/TagCLIP} \\
        \hline
    \end{tabular}
\end{table}

\begin{table}[t]
    \centering
    \caption{Details of the datasets used in the study.}
    \label{tab:dataset-details}
    \begin{tabular}{|p{4cm}|p{10cm}|}  
        \hline
        \textbf{Dataset} & \textbf{Link} \\
        \hline
        COCO 2014 val~\cite{lin2014microsoft} & \url{http://images.cocodataset.org/zips/val2014.zip} \\
        M-HalDetect~\cite{gunjal2024detecting} & \url{https://github.com/hendryx-scale/mhal-detect} \\
        \hline
    \end{tabular}
\end{table}
\section{Model and data}\label{appendix-models}
We include the model and data details in Table~\ref{tab:model-data}.

\section{Prompt for extracting objects from image captions using LLM}\label{appendix-prompt}
To extract explicitly mentioned objects from image captions using Llama-3.1-8B~\cite{dubey2024llama}, use the following prompt template:

template = """You will be provided with an image caption. For example:
"The image depicts a large kitchen with a man and a woman preparing food. The man is standing in front of the stove, while the woman is closer to the refrigerator. There are several appliances in the kitchen, including an oven, a microwave, and a dishwasher. A clock can be seen hanging on the wall, adding a sense of time to the scene. Various utensils are scattered around the kitchen, such as knives, forks, and spoons, suggesting that they are being used for food preparation. In addition, there are two cups visible in the scene, one close to the man and the other near the woman. Overall, the kitchen appears to be bustling with activity as the couple works together to prepare their meal."

Extract all explicitly mentioned objects (nouns that clearly exist in the scene) from the caption. Exclude objects that are described with uncertainty (e.g., "might be," "possibly"). List the objects in singular form, regardless of whether they appear as singular or plural in the caption. Do not include any attributes or descriptions of the objects. Separate each object with ". " (period and space).

For the given example, the result would be the following:
kitchen. man. woman. stove. refrigerator. appliance. oven. microwave. dishwasher. clock. utensil. knife. fork. spoon. cup.

Here is the caption:
"{caption}"

Please only output extracted objects:
"""

\section{Dataset construction and model training for object hallucination detection Pipeline}\label{appendix-pipeline}

To determine optimal weights for combining confidence scores from YOLO-World, Grounding DINO, and TagCLIP, we train a logistic regression model using a constructed dataset.

\subsection*{Dataset construction}
As there is no existing dataset specifically annotated for object hallucinations in image captions, we re-utilize the M-HalDetect dataset (introduced in Section~\ref{subsec-patterns}). We extract objects from captions using the LLM-based approach described in Section~\ref{subsec-pipeline}. Object labels are assigned based on M-HalDetect annotations using the following criteria:
\begin{itemize}
    \item If an object appears in a text segment labeled as \textbf{ACCURATE}, it is assigned an ACCURATE label.
    \item If an object appears in a text segment labeled as \textbf{INACCURATE}, it is assigned an INACCURATE label.
    \item If an object is only mentioned in other categories, it remains unlabeled.
\end{itemize}

It's important to note that since M-HalDetect was originally designed to capture all forms of hallucination, not exclusively object presence, this labeling approach may not provide completely accurate labels for object hallucinations.

\subsection*{Data balancing and model training}
The initial dataset contains substantially more ACCURATE labels. We address this imbalance through downsampling, creating a balanced dataset for logistic regression model training. The model parameters are as follows:

\begin{itemize}
    \item \textbf{Model Architecture} \\
          LogisticRegression(max\_iter=1000)
    \item \textbf{Original Data Distribution} \\
          Label 0: 2313, Label 1: 10793
    \item \textbf{Downsampled Data} \\
          4626 samples (2313 per class)
    \item \textbf{Features Used} \\
          YOLO\_conf, DINO\_conf, TagCLIP\_conf
    \item \textbf{Cross-Validation} \\
          5-fold K-Fold Cross-Validation
    \item \textbf{Model Output Formula} \\
          \[
          \text{logit}(p) = \text{intercept} + \text{coef1} \times \text{YOLO\_conf}
          + \text{coef2} \times \text{DINO\_conf} + \text{coef3} \times \text{TagCLIP\_conf}
          \]
\end{itemize}

Performance comparison of different feature combinations is presented in Table ~\ref{tab:performance}.

\begin{table}
    \centering
    \caption{Performance comparison of different feature combinations}
    \label{tab:performance}
    \begin{tabular}{lccc}
        \toprule
        \textbf{Model} & \textbf{Accuracy} & \textbf{Precision} & \textbf{Recall} \\
        \midrule
        YOLO only & 0.7216 ± 0.0117 & 0.7534 ± 0.0147 & 0.6597 ± 0.0214 \\
        DINO only & 0.7041 ± 0.0181 & 0.6835 ± 0.0165 & 0.7608 ± 0.0178 \\
        TagCLIP only & 0.6857 ± 0.0190 & 0.7404 ± 0.0161 & 0.5719 ± 0.0324 \\
        Combined (YOLO+DINO+TagCLIP) & 0.7453 ± 0.0116 & 0.7621 ± 0.0120 & 0.7142 ± 0.0191 \\
        \bottomrule
    \end{tabular}
\end{table}

\section{Training of token-level hallucination classifier}\label{appendix-trainingclassifier}

To accurately detect token-level hallucinations in image captions, we developed and trained a Multi-Layer Perceptron (MLP) classifier. In the following, we provide a detailed overview of the model architecture, training process, and the parameters used.

\subsection*{Model Architecture and Training Parameters}

\begin{itemize}
    \item \textbf{Model Architecture}:
    MLP with hidden layers: (256, 128, 64). Each hidden layer consists of: Linear $\rightarrow$ BatchNorm1d $\rightarrow$ ReLU $\rightarrow$ Dropout(0.5). The final layer is Linear $\rightarrow$ 1 output dimension.
    
    \item \textbf{Loss Function}:
    BCEWithLogitsLoss. Combines Sigmoid activation and binary cross-entropy loss into a single function for stability.
    
    \item \textbf{Optimizer}:
    AdamW. Adam optimizer with decoupled weight decay to mitigate overfitting.
    
    \item \textbf{Initial Learning Rate}:
    $1 \times 10^{-3}$. The starting learning rate for the optimization process.
    
    \item \textbf{Weight Decay}:
    $1 \times 10^{-5}$. L2 regularization factor used in AdamW to control model complexity.
    
    \item \textbf{Batch Size}:
    512. Number of samples processed before the model's internal parameters are updated.
    
    \item \textbf{Number of Epochs}:
    40. Total training epochs per fold.
    
    \item \textbf{Dropout Probability}:
    0.5. Applied to each hidden layer to reduce overfitting by randomly deactivating neurons during training.
    
    \item \textbf{Learning Rate Scheduler}:
    ReduceLROnPlateau. Reduces the learning rate by a factor of 0.1 when the validation loss plateaus for 3 consecutive epochs.
    
    \item \textbf{Train/Val/Test Split}:
    0.6 / 0.2 / 0.2. The dataset is divided into 60\% training, 20\% validation, and 20\% testing subsets.
    
    \item \textbf{Cross-Validation}:
    KFold, $n=5$. Utilizes 5-fold cross-validation on the training set to evaluate model performance and ensure robustness.
    
    \item \textbf{Ensemble Method}:
    Averaging predicted probabilities. The final prediction is obtained by averaging Sigmoid-transformed logits across folds and applying a threshold of 0.5.

    \item \textbf{Training Set Size}:
    124,234.
    
    \item \textbf{Validation Set Size}:
    41,411. 
    
    \item \textbf{Test Set Size}:
    41,413.
    
    \item \textbf{Total Dataset Size}:
    207,058.
    
    \item \textbf{Class 0 (INACCURATE)}:
    55,574 (26.84\%). 
    
    \item \textbf{Class 1 (ACCURATE)}:
    151,484 (73.16\%).
\end{itemize}

\section{Proposed Method: Caption Example}\label{appendix-captionsample}

\begin{figure}[t]
    \centering
    \includegraphics[width=0.8\textwidth]{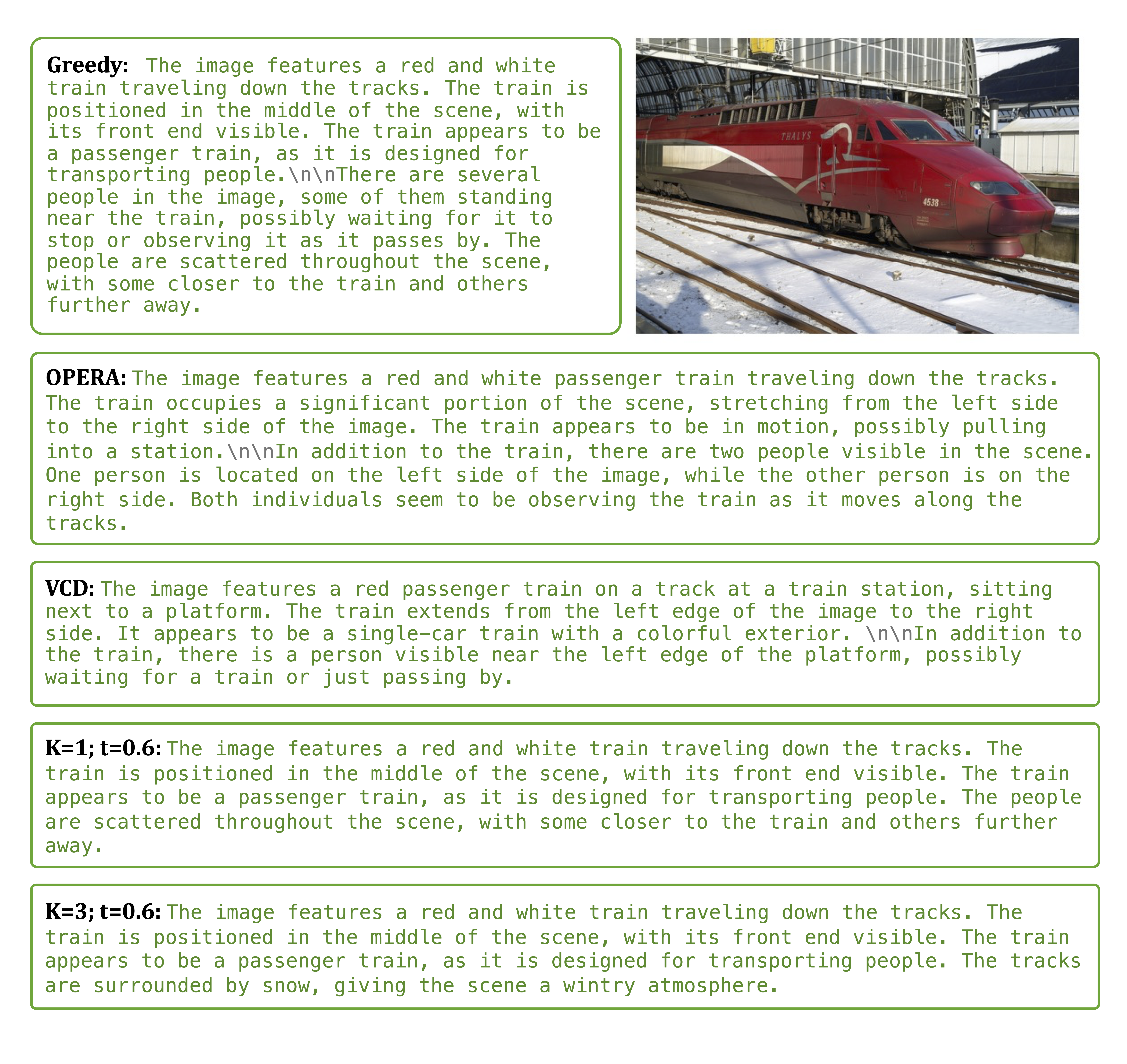}
    \caption{Captions sampled by different decoding methods.}
    \label{fig:caption samples}
\end{figure}

Figure~\ref{fig:caption samples} shows captions sampled by different decoding methods.

\end{document}